\documentclass{article}
\usepackage{spconf,graphicx}
\usepackage{graphicx}
\usepackage[linesnumbered,ruled,vlined]{algorithm2e}
\SetKwInput{KwInput}{Input}                
\SetKwInput{KwOutput}{Output}              
\SetKwInput{KwReturn}{Return} 
\usepackage[justification=centering]{caption}
\usepackage{algorithmic}
\usepackage{amssymb}
\usepackage{amsmath}
\usepackage{multirow}
\usepackage{adjustbox}
\usepackage{caption}
\usepackage{hyperref}

\title{Dropout Concrete Autoencoder for Band Selection on HSI Scenes}
\name{Lei Xu, Mete Ahishali\thanks{This work has been supported by the NSF-Business Finland project AMALIA.}, and Moncef Gabbouj}
\address{Faculty of Information Technology and Communication Sciences, \\
Tampere University, Tampere, Finland}

\begin{document}
%
\maketitle
\begin{abstract}
Deep learning-based informative band selection methods on hyperspectral images (HSI) recently have gained intense attention to eliminate spectral correlation and redundancies. However, the existing deep learning-based methods either need additional post-processing strategies to select the descriptive bands or optimize the model indirectly, due to the parameterization inability of discrete variables for the selection procedure. To overcome these limitations, this work proposes a novel end-to-end network for informative band selection. The proposed network is inspired by the advances in concrete autoencoder (CAE) and dropout feature ranking strategy. Different from the traditional deep learning-based methods, the proposed network is trained directly given the required band subset eliminating the need for further post-processing. Experimental results on four HSI scenes show that the proposed dropout CAE achieves substantial and effective performance levels outperforming the competing methods. The code is available at \url{https://github.com/LeiXuAI/Hyperspectral}.
\end{abstract}
\begin{keywords}
band selection, concrete autoencoder, dropout feature ranking, hyperspectral image data
\end{keywords}
\section{Introduction}
\label{sec:intro}

Hyperspectral images (HSI) captured by hyperspectral remote sensing imaging spectrometers \cite{Ngadi2010HyperspectralTechniques} cover a wide and continuous range of the electromagnetic spectrum beyond the visible wavelengths with multiple spectral bands. Due to this characteristic, hyperspectral images contain enormous information utilizing its various applications, such as in precision agriculture \cite{Bajwa2004HyperspectralApplications}, mineral detection \cite{Siebels2020EstimationApproach}, and classification of landscape \cite{Camps-Valls2014AdvancesMethods}. However, massive spectral bands of hyperspectral images imply information redundancy which leads to “Hughes
phenomenon” \cite{Feng2021ConvolutionalSelection}, computational complexity, and larger storage capacity \cite{Morales2021HyperspectralNetworks}. Considering the unattainability of selecting prominent wavelengths before capture \cite{Morales2021HyperspectralNetworks}, it is indispensable to develop band selection algorithms in advance for the subsequent tasks with hyperspectral images. 

In coping with the band redundancy problem existing in hyperspectral images, various band selection methods have been proposed, such as ranking-based methods \cite{Chang1999AClassification, Koonsanit2012BandTechnique}, clustering-based methods \cite{Sun2015BandClassification, Wang2018OptimalSelection}, searching-based methods \cite{Morales2021HyperspectralNetworks}, etc. Ranking-based methods are unsupervised approaches exploring specific criteria to rank frequency bands based on the distinctive information of each band. Principal components analysis (PCA) is generally used to extract descriptive features through projection in an optimal lower-dimensional space. As demonstrated in \cite{Chang1999AClassification, Koonsanit2012BandTechnique}, band prioritization is based on the eigen (spectral) decomposition. Finally, clustering-based methods \cite{Wang2018OptimalSelection, Wang2021ASelection} usually try to group relevant bands in selected subsets by computing similarity matrices. Sun et al. proposed an improved sparse subspace clustering (ISSC) method in \cite{Sun2015BandClassification} to select informative bands with an angular-based similarity matrix. These search-based methods utilize specific search strategies to find the optimal subset with the most informative bands. Similarly, in \cite{Wang2021ASelection}, Wang et al. proposed a fast neighborhood grouping method to explore the context information of the spectral information for the informative partitioned group selection with a coarse-to-fine mechanism. 

Various deep learning-based methods have been explored to address the band redundancy problem as shown in \cite{Feng2021ConvolutionalSelection, Morales2021HyperspectralNetworks, Cai2020BS-Nets:Image, Roy2021DARecNet-BS:Selection, Ahishali2022SRL-SOA:SELECTION}. In \cite{Roy2021DARecNet-BS:Selection}, an end-to-end unsupervised network was proposed for band selection, which employs a dual-attention mechanism, i.e., a position attention module and a channel attention module. The dual attention mechanism can capture the long-range nonlinear interdependencies from the spectral and spatial directions. Feng et al. proposed another end-to-end unsupervised convolutional neural network combining band selection, feature extraction, and classification. The proposed network uses a hard thresholding strategy to constrain the weights of convolution kernels and select spectral bands after band-wise independent convolution. A straight-through estimator and a coarse-to-fine loss are introduced to obtain optimal weights. 

Moreover, studies in \cite{Feng2021ConvolutionalSelection, Cai2020BS-Nets:Image, Ahishali2022SRL-SOA:SELECTION, Liu2022AImage} have proposed various Autoencoder (AE) models for the band selection task, where after HSI reconstruction with the selected bands, a classifier, such as SVM or KNN, is used for classification and the final evaluation is performed on the classification results. Cai et al. proposed an end-to-end AE-based framework, BS Network \cite{Cai2020BS-Nets:Image}, for band selection based on fully connected networks (BS-Net-FC) and convolutional networks (BS-Net-Conv) with an attention mechanism. The BS network explores a band attention module to explicitly model nonlinear interdependencies between the spectral bands \cite{Cai2020BS-Nets:Image} with learned weights. Finally, a reconstruction network restores the original spectral bands with the selected number of the reweighted bands. Ahishali et al. proposed another AE-based band selection model, Self-Representation Learning with Sparse 1D-Operational Autoencoder (SRL-SOA) in \cite{Ahishali2022SRL-SOA:SELECTION}. The SRL-SOA model consists of a single 1D-operational layer encoder with generative neurons for mapping and a self-representation pixel-wise decoder for reconstruction. The generative neurons use Taylor series expansion with trainable parameters for band selection in the representation matrix.

Concrete autoencoder (CAE), as an embedded unsupervised feature selection method, has been explored for various feature selection \cite{Abid2019ConcreteReconstruction} tasks. The CAE inspired by the concrete distribution \cite{Maddison2017TheVariables, Jang2017CategoricalGumbel-softmax} aims to learn an informative subset and reconstruct the input data from this subset simultaneously. The concrete distribution is introduced for reparameterizations of discrete random variables as of continuous random variables while optimizing stochastic computation graph via gradient descent \cite{Maddison2017TheVariables}. For instance, Sun et al. proposed a Gumbel-softmax-based CAE and an information entropy criterion for optimal band subset selection in \cite{Sun2022NovelSelection}. The Gumble-softmax distribution \cite{Jang2017CategoricalGumbel-softmax} can transform a discrete weight matrix into continuous variables for the selected subset optimization during the backpropagation for local optimal solutions. Finally, the information entropy criterion searches a global optimal band subset. In \cite{Sun2022StochasticSelection}, a novel stochastic gate was proposed as a differential layer in AE-based architecture for parameterization process based on a Gaussian-based relaxation of Bernoulli variables. The Stochastic gate is learnable for an optimal band subset selection without post-processing for a global optimal result. Although the methods mentioned above have achieved certain outcomes, common limitations still exist in these works. For instance, the nonlinear relation of bands lacks investigation \cite{Ahishali2022SRL-SOA:SELECTION} due to the linear convolutional operations as in \cite{Feng2021ConvolutionalSelection, Sun2022NovelSelection}. In addition, the required band subset is not optimized directly by the model but using an approximation of learned weights ranking \cite{Cai2020BS-Nets:Image} or band entropy ranking \cite{Roy2021DARecNet-BS:Selection}. 

In this work, we propose a novel CAE method based on a dropout feature ranking strategy for HSI band selection task without any post-processing step. The architecture of the proposed method consists of a concrete selector layer for informative subset selection and a standard decoder part for HSI reconstruction. It utilizes the Binary Concrete relaxation \cite{Maddison2017TheVariables} and dropout feature ranking (Dropout FR) strategy \cite{Chang2017DropoutModels} to learn the nonlinear dependencies of spectral bands with the concrete selector layer. Moreover, the model is optimized directly and converges to a fixed optimal subset under the same condition during training. 

The remainder of this work is organized as follows. Section 2 provides the theories of the proposed Dropout CAE in detail. Section 3 presents the datasets, implementation details, and experimental results. Finally, we conclude this work in Section 4. 

\section{Proposed method}
\label{sec:pro_method}
In this section, we first present the principle of concrete distribution, dropout feature ranking, and the proposed Dropout CAE in detail. Next, the pseudo-code of the proposed method is provided at the end of this section.

\subsection{Concrete Distribution}
The concrete random variables are defined as a continuous relaxation of discrete random variables \cite{Maddison2017TheVariables, Jang2017CategoricalGumbel-softmax}, which are introduced to address the parameterization inability of discrete random variables during the loss propagation by gradient descent. The construction of the concrete random variables is motivated by the Gumbel-Max trick \cite{Maddison2014ASampling} sampling from a discrete distribution with \textit{argmax}. The discrete distribution \cite{Maddison2017TheVariables} is depicted as one-hot vectors $d \in \{0, 1\}^n$ and $\sum_{k=1}^{n} d_k = 1$. The Gumbel-Max trick cannot be directly used for gradient descent because the \textit{argmax} is a non-differentiable operation. Then the softmax function is introduced to replace the \textit{argmax} for a continuous relaxation of one-hot vector. 

The Gumbel-Softmax distribution as a novel concrete distribution has a closed-form density on the simplex defined with location parameters $\alpha \in (0, \infty)^n$ and a temperature parameter $\tau \in (0, \infty)$ \cite{Maddison2017TheVariables}. $X \sim \text{Concrete}(\alpha, \tau)$ depicts  $X$ has the concrete distribution. Then each element $X_{k}$ is sampled as

\begin{equation}
\label{eq:gumbel_softmax}
X_{k} = \frac{\text{exp}((\log \alpha_k + G_k)/\tau)}{\sum_{i=1}^{n} \text{exp}((\log \alpha_i + G_i)/\tau)}, 
\end{equation}
where $G_k \sim \text{Gumbel} \: i.i.d$. Such computation achieves a random probability vector summing to 1. As the temperature parameter $\tau \rightarrow 0$, $X_{k}$ is smoothly annealed to the computation of the discrete \textit{argmax}, which means the Gumbel-Softmax distribution can obtain near one-hot samples with a proper temperature $\tau$ setting schedule \cite{Jang2017CategoricalGumbel-softmax}. 

The encoder part of the CAE architecture consists of a concrete selector layer based on the concrete distribution. Accordingly, the selector layer samples a concrete random variable using a temperature parameter $\tau \in (0, \infty)$ and parameters $\alpha_k \in (0, \infty)$ for a continuous relaxation of a one-hot vector \cite{Abid2019ConcreteReconstruction, Maddison2017TheVariables}.

\subsection{Dropout Feature Ranking}

Bernoulli distribution is a special case of the Gumbel-Max trick with two states of the discrete random variable on $\{0, 1\}^2$. When the Gumbel-Softmax trick is implemented on the Bernoulli distribution for a Binary Concrete random variable, the sampled element $X \in (0, 1)$ \cite{Maddison2017TheVariables} is depicted as
\begin{equation}
\label{eq:gumbel_bin}
X = \frac{1}{1 + \text{exp}(-(\log \alpha + L)/\tau)}, 
\end{equation}
where $L \sim Logistic $. When the temperature parameter $\tau$ follows a proper schedule approach to 0, the output is $X=1$ with the probability of $\alpha/(1+\alpha)$.

Variational Dropout as a regularization technique is initially proposed to solve the overfitting problem in deep learning models \cite{Srivastava2014Dropout:Overfitting}. The Bernoulli distribution is utilized in the variation dropout strategy as a dropout mask in deep learning models. Dropout masks can stochastically determine whether the hidden node in a layer is retained or dropped \cite{Chang2017DropoutModels} using a learnable dropout rate indicating the feature importance, i.e., a smaller dropout rate means more representative feature. The proposed Dropout FR loss function is defined as
\begin{equation}
\label{eq:drop_out}
\mathcal{L}(\theta) = -\frac{1}{N} \sum_{i=1}^{N} \log p(\mathbf{y_i}| f_\theta(\mathbf{x}_i \odot \mathbf{m}_i)) + \frac{\lambda}{N} \sum_{i=1}^{N} \sum_{j=1}^{D} m_{ij},  
\end{equation}
where $\mathbf{m}_i \sim {q}_{\mathbf{\theta}}(\mathbf{m})$; ${q}_{\theta}(\mathbf{m})$ is a variational mask distribution defined as ${q}_{\theta}(\mathbf{m}) = \Pi_{j=1}^{D} q(m_j|\theta_j) = \Pi_{j=1}^{D} \text{Bern}(m_j|\theta_j)$ and $\theta_j$ is the dropout rate of feature j. $N$ is the batch size and $\lambda$ is the regularization hyperparameter.

\subsection{Dropout Concrete Autoencoder}
\begin{figure*}[ht]
  \centering
  \includegraphics[width=4.0in]{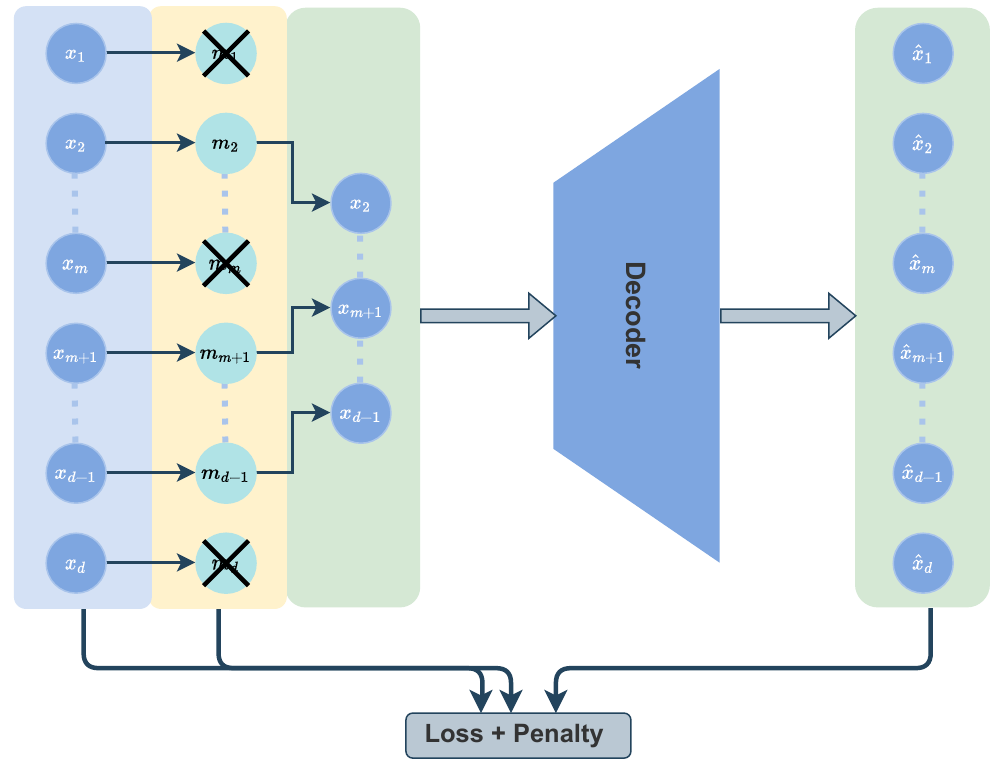}
  \caption{The architecture of Dropout CAE with the proposed concrete selector layer.}
  \label{fig:schema}
\end{figure*}

The proposed Dropout CAE aims to select the $\textit{k}$ number of most informative spectral bands as a subset with the encoder part. To achieve this objective, we propose a concrete selector layer in the encoder consisting of binary concrete distribution \cite{Maddison2017TheVariables} integrated with the variational dropout feature ranking strategy \cite{Chang2017DropoutModels}. The decoder part of the network has two fully connected layers performing unsupervised reconstruction, i.e., $f_\theta(\cdot)$ using the selected subset. The overall architecture of the proposed network is shown in Fig. \ref{fig:schema}. 

The concrete selector layer takes a 2D HSI matrix $\mathbf{x} \in \mathbb{R}^{N \times d}$ as input, where $N$ is the number of pixels and $d$ is the number of spectral bands. The concrete selector layer samples $d$-dimensional Binary Concrete random variables $\mathbf{m} \in \mathbb{R}^{d \times 1}, \mathbf{m} \in \{0, 1 \}^2$ using Eq.~\eqref{eq:gumbel_bin}. The output of the concrete selector layer is computed by $\mathbf{x} \odot \mathbf{m}$, which merely retains $\textit{k}$ number of spectral bands as the input of the decoder. The reconstruction performed by the decoder is formulated as 
\begin{equation}
\label{eq:recon_x}
\mathbf{\hat{x}} = f_\theta(\mathbf{x} \odot \mathbf{m}),
\end{equation}
where $\mathbf{\hat{x}} \in \mathbb{R}^{N \times d}$. The variational mask $\mathbf{m}$ is obtained by optimization of the Dropout FR loss function w.r.t. the dropout rate $\theta$ as
\begin{equation}
\label{eq:drop_cae}
\mathcal{L}(\theta) = -\frac{1}{N} \sum_{i=1}^{N} \sum_{j=1}^{d} x_{ij}\log (\hat{x}_{ij}) + \frac{\lambda}{N} \sum_{i=1}^{N} \sum_{j=1}^{d} m_{ij},  
\end{equation}
where $\lambda$ is a hyperparameter for the regularization and $\hat{x}_{ij}$ is reconstructed HSI pixel. During the training phase, the temperature parameter $\tau$ gets infinitely close to $0$, and according to Eq. ~\eqref{eq:gumbel_bin}, the sampled element $X$ values become $1$ retaining $\textit{k}$ nodes with the lowest dropout rates.

\begin{algorithm}
\SetAlgoLined
\tcc{Training procedure for obtaining an optimal band subset .}
\KwInput{2D HSI matrix $\mathbf{x} \in \mathbb{R}^{N \times d}$,\:the desired band number $\textit{k}$,\:the temperature parameter $\tau$,\:learnable dropout rate $\theta$,\:regularization $\lambda$,\:number of epochs C.}
\KwOutput{Indices of the selected $\textit{k}$ number of bands.}
 Pre-process the dataset for training\;
 Initialize $\theta$, $\tau$\;
 \For{$c \in \{1,\dots,C\}$}{
  Obtain the output $\mathbf{x} \odot \mathbf{m}$ using the concrete selector layer\;
  Get the required $\textit{k}$ number of bands, according the lowest $\textit{k}$ learning rate $\theta$\;
  Reconstruct the input HSI with the selected $\textit{k}$ spectral bands as $\mathbf{\hat{x}}$ using the encoder as Eq.~\eqref{eq:recon_x}\; 
  Compute the updated model weights by ADAM optimizer minimizing the loss function Eq.~\eqref{eq:drop_cae}\;
  Adjust the temperature parameter $\tau$ with a fixed schedule\; }
\KwReturn{decoder $f_\theta(\cdot)$ and binary concrete parameters}
 \caption{The pseudo-code of the proposed Dropout CAE for band selection}
\end{algorithm}

\section{Experimental Evaluation}
In this section, we first describe the datasets used in the experiments, and we provide implementation details which are followed by comparative evaluations. 

\subsection{Datasets}
In this work, we have evaluated the proposed network using four HSI scenes: Indian Pines \cite{Baumgardner20152203}, PaviaU \cite{GambaPaviaUniversity}, Salinas \cite{EUDAT2014Salinas}, and KSC \cite{Bert2016AmeriFluxOak}. The original HSI scenes contain two spatial dimensions and one spectral dimension denoted as $\mathbf{x}_{ori} \in \mathbb{R}^{h \times w \times d}$. Then each original HSI scene is transformed into a 2D HSI matrix $\mathbf{x} \in \mathbb{R}^{N \times d}$ for training. $N$ is the number of pixels for training and $d$ is the number of spectral bands. The pixel values of $\mathbf{x} \in \mathbb{R}^{N \times d}$ are normalized to the range [0, 1].

\begin{itemize}
    \item \textit{Indian Pines scene} \cite{Baumgardner20152203} is captured by the AVIRIS sensor with $145 \times 145$ pixels and 224 spectral bands in the wavelength range from 0.4 $\mu$m to 2.5 $\mu$m. The Indian Pines scene contains dynamic geographical features and man-made facilities, such as agriculture, forest, lane highways, etc. The number of bands for the experiment is 200 after removing bands covering the water absorption region:[104-108], [150-163], 220. The number of pixels for training from this scene is 10249.
    \item \textit{PaviaU scene} \cite{GambaPaviaUniversity} is captured by the ROSIS sensor with $610 \times 610$ pixels and 103 spectral bands. It has a 1.3-meter resolution and 9 classes with 42776 number of training pixels.
    \item \textit{Salinas scene} \cite{EUDAT2014Salinas} is captured by the AVIRIS sensors with a high spatial resolution (3.7-meter pixels), which contains 512 lines by 217 samples. This scene contains 16 classes, such as bare soils, vegetables, etc. After 20 water absorption bands: [108-112], [154-167], 224 are discarded, and 204 bands are used for this work. There is 54129 pixels available for training.
    \item \textit{KSC scene} \cite{Bert2016AmeriFluxOak} is captured by the AVIRIS sensor with a spatial resolution of 18 m. It contains 224 bands of 10 nm width with center wavelengths from 400 - 2500 nm and 5211 pixels for training.
\end{itemize}

\begin{table*}[!ht]
    \centering
    \begin{adjustbox}{width=\textwidth,center}
    \begin{tabular}{c|ccc|ccc|ccc|ccc}
    \hline
        \multirow{2}{*}{Dataset} & \multicolumn{3}{c|}{Indian Pines \: (25 bands)} & \multicolumn{3}{c|}{Salinas \: (20 bands)} & \multicolumn{3}{c|}{PaviaU \: (15 bands)} &\multicolumn{3}{c}{KSC \: (15 bands)}\\ 
        & OA & AA & Kappa & OA & AA & Kappa & OA & AA & Kappa & OA & AA & Kappa \\ \hline
        SRL-SOA Q=1 & 0.8070 & 0.7662 & 0.7793 & 0.9280 & \textbf{0.9640} & 0.9197 & 0.9106 & 0.8838 & 0.8807 & 0.8664 & 0.8032 & 0.8511 \\ 
        SRL-SOA Q=3 & 0.7779 & 0.7433 & 0.7448 & 0.9281 & 0.9627 & \textbf{0.9198} & \textbf{0.9190} & 0.8912 & 0.8920 & 0.8674 & 0.8059 & 0.8521 \\ 
        SRL-SOA Q=5 & 0.7972 & 0.7617 & 0.7678 & 0.9258 & 0.9602 & 0.9173 & \textbf{0.9181} & 0.8897 & 0.8908 & \textbf{0.8878} & 0.8308 & \textbf{0.8750} \\ 
        SpaBS & 0.6298 & 0.5400 & 0.5688 & 0.9032 & 0.9359 & 0.8919 & 0.8526 & 0.8126 & 0.7998 & 0.8391 & 0.7642 & 0.8205 \\ 
        EGCSR & \textbf{0.8072} & 0.7737 & \textbf{0.7795} & 0.9215 & 0.9592 & 0.9125 & 0.8624 & 0.8345 & 0.8143 & 0.8803 & 0.8175 & 0.8666 \\ 
        ISSC & \textbf{0.8123} & 0.7643 & \textbf{0.7853} & 0.9304 & \textbf{0.9661} & \textbf{0.9224} & 0.9149 & 0.8891 & 0.8865 & \textbf{0.8920} & 0.8373 & \textbf{0.8797} \\
        BS-Net-FC & 0.6945 &0.7439 & 0.7080 & 0.9551 & 0.9174 & 0.9080 & 0.8761 & 0.9066 & 0.8760 & 0.8045 & \textbf{0.8765} & 0.8620 \\ 
      \hline
        Dropout CAE ($T_1$) & 0.7701 & \textbf{0.7786} & 0.7480 & \textbf{0.9611} & 0.9252 & 0.9166 & 0.8949 & \textbf{0.9215} & \textbf{0.8955} & 0.8003 & 0.8701 & 0.8553 \\ 
        Dropout CAE ($T_2$) & 0.7600 & \textbf{0.7827} & 0.7527 & \textbf{0.9614} & 0.9271 & 0.9187 & 0.8991 & \textbf{0.9226} & \textbf{0.8970} & 0.8100 & \textbf{0.8753} & 0.8610 \\ 
        Dropout CAE ($T_3$) & 0.7146 & 0.7629 & 0.7300 & 0.9570 & 0.9170 & 0.9074 & 0.8865 & 0.9115 & 0.8820 & 0.7628 & 0.8381 & 0.8196 \\ 
        
        \hline
          All Bands & 0.7965 & 0.7244 & 0.7670 & 0.9342 & 0.9663 & 0.9266 & 0.9438 & 0.9234 & 0.9252 & 0.9127 & 0.8677 & 0.9027 \\ 
          \hline
    \end{tabular}
    \end{adjustbox}
     \caption{Overall comparison of the competing and proposed Dropout CAE band selection methods evaluated on four datasets.}
     \label{tab:results}
\end{table*}
\subsection{Implementation} 
\textbf{Settings:} The proposed Dropout CAE is implemented with PyTorch \cite{Paszke2019PyTorch:Library} on a Nvidia GPU cluster platform. The number of hidden neurons in the decoder part is 128. The hyperparameters for training the model are set as follows: the optimizer used in this work is ADAM, the learning rate is 0.001, $\beta_1 = 0.9$, $\beta_2 = 0.999$. We adopt the multistep learning rate strategy in PyTorch \cite{Paszke2019PyTorch:Library}, by which two milestones 15 and 30 are set with $\gamma=0.1$. The $\lambda = 0.005$ in Eq.~\eqref{eq:drop_cae}. The corresponding hyperparameters of competing methods are set to the default values. 

In the experiments, same samples that are annotated for the classification are used to train the proposed Dropout CAE model. Then support vector machine (SVM) \cite{Cai2020BS-Nets:Image, Melgani2004ClassificationSensing} classifier is used for the performance evaluation with the selected band subset. We randomly select $10\%$ of annotated samples from each data scene for training the SVM classifier and $90\%$ of samples for testing. We run the classification process ten times on each data scene independently for a more fair and robust comparison.

\textbf{Evaluation Metrics:} We adopt three quantitative metrics: overall accuracy (OA), average accuracy (AA), and kappa coefficient (Kappa) \cite{Cai2020BS-Nets:Image, Ahishali2022SRL-SOA:SELECTION} for the classification performance of the selected band subsets to evaluate the effectiveness of the Dropout CAE. The final evaluation results are averaged from the ten runs. Besides, we adopt entropy to measure the information content of each band in each scene, and then make a comparison with the selected band distribution \cite{Cai2020BS-Nets:Image} from a sample run. 

\textbf{Annealing Schedule:} The Dropout CAE model is optimized with regard to the dropout rate $\theta$, which is highly affected by the setting of the temperature parameter $\tau$. Regardless of whether $\tau$ is high or low, the concrete selector layer converges to a sub-optimal informative band subset with a fixed temperature. Therefore, the temperature parameter $\tau$ is set as a schedule that can gradually decay at each epoch according to a first-order exponential decay as \cite{Abid2019ConcreteReconstruction}
\begin{equation}
\label{eq:temperature_sche}
\tau = \tau_0 (\tau_C / \tau_0)^{B/(N \times C)},  
\end{equation}
where $\tau_0$ is the start temperature holding a higher value, $\tau_C$ is the final temperature with a lower value, $\textit{C}$ is the total number of epochs, $\textit{B}$ is the batch size, and $\textit{N}$ is the total number pixels for each scene. The annealing schedule begins with the $\tau_0$ and smoothly decays to $\tau_C$. In this work, we set three kinds of annealing schedule parameters as an ablation study to investigate the effect of the annealing schedule on the results. The first setting ($T_1$) is $\tau_0=1$, $\tau_C=0.001$, $\textit{C}=40$, $\textit{B}=1$. The second one ($T_2$) is $\tau_0=1$, $\tau_C=0.001$, $\textit{C}=200$, $\textit{B}=256$. The last one ($T_3$) is $\tau_0=1$, $\tau_C=0.01$, $\textit{C}=200$, $\textit{B}=32$.
\subsection{Comparisons}

\begin{figure}[ht]
  \centering
  \includegraphics[width=1.1\columnwidth]{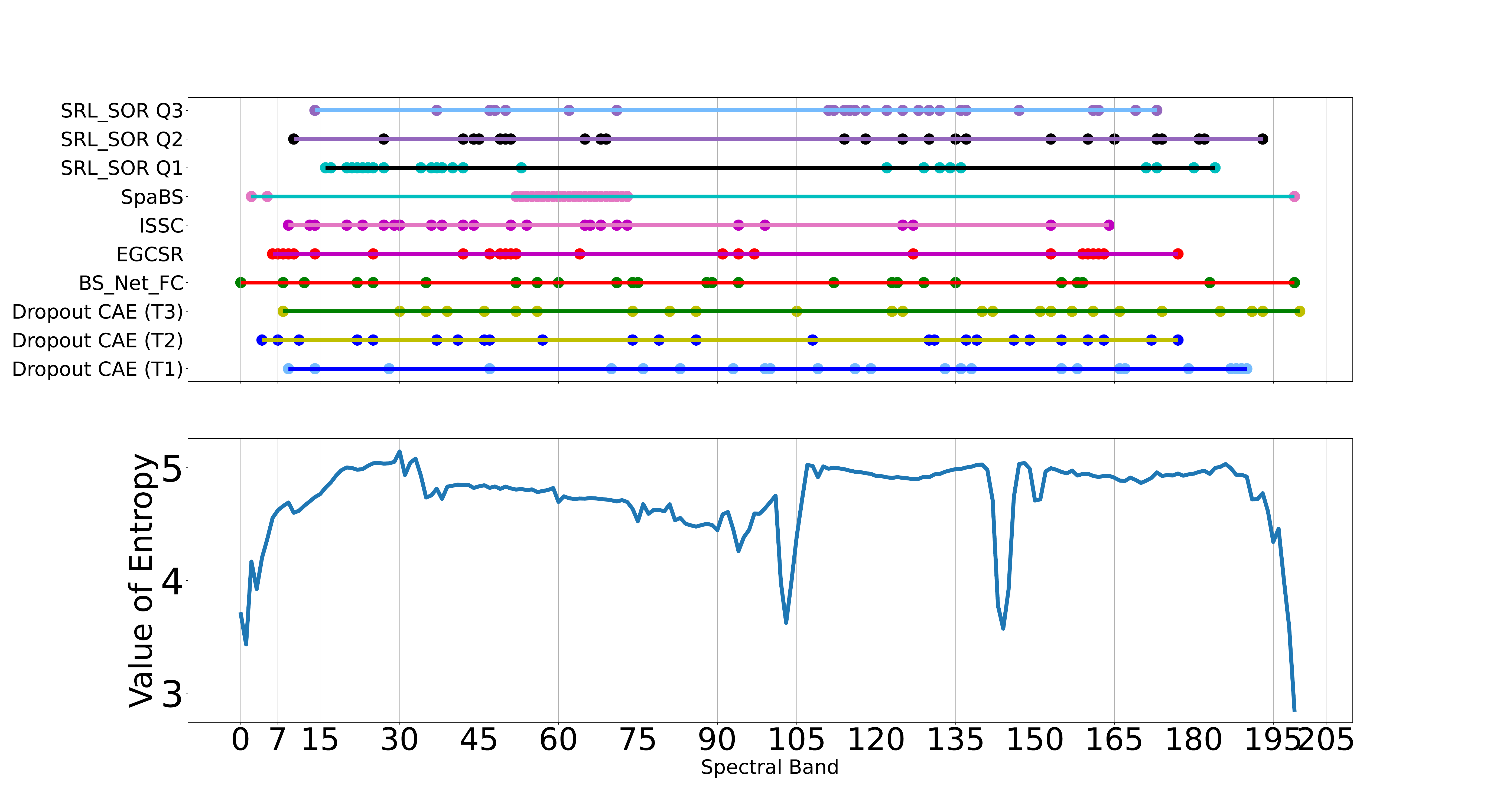}
  \caption{The selected subset with 25 bands and entropy values of each band on the Indian Pines scene.}
  \label{fig:indian_entropy}
\end{figure}

\begin{figure}[ht]
  \centering
  \includegraphics[width=1.1\columnwidth]{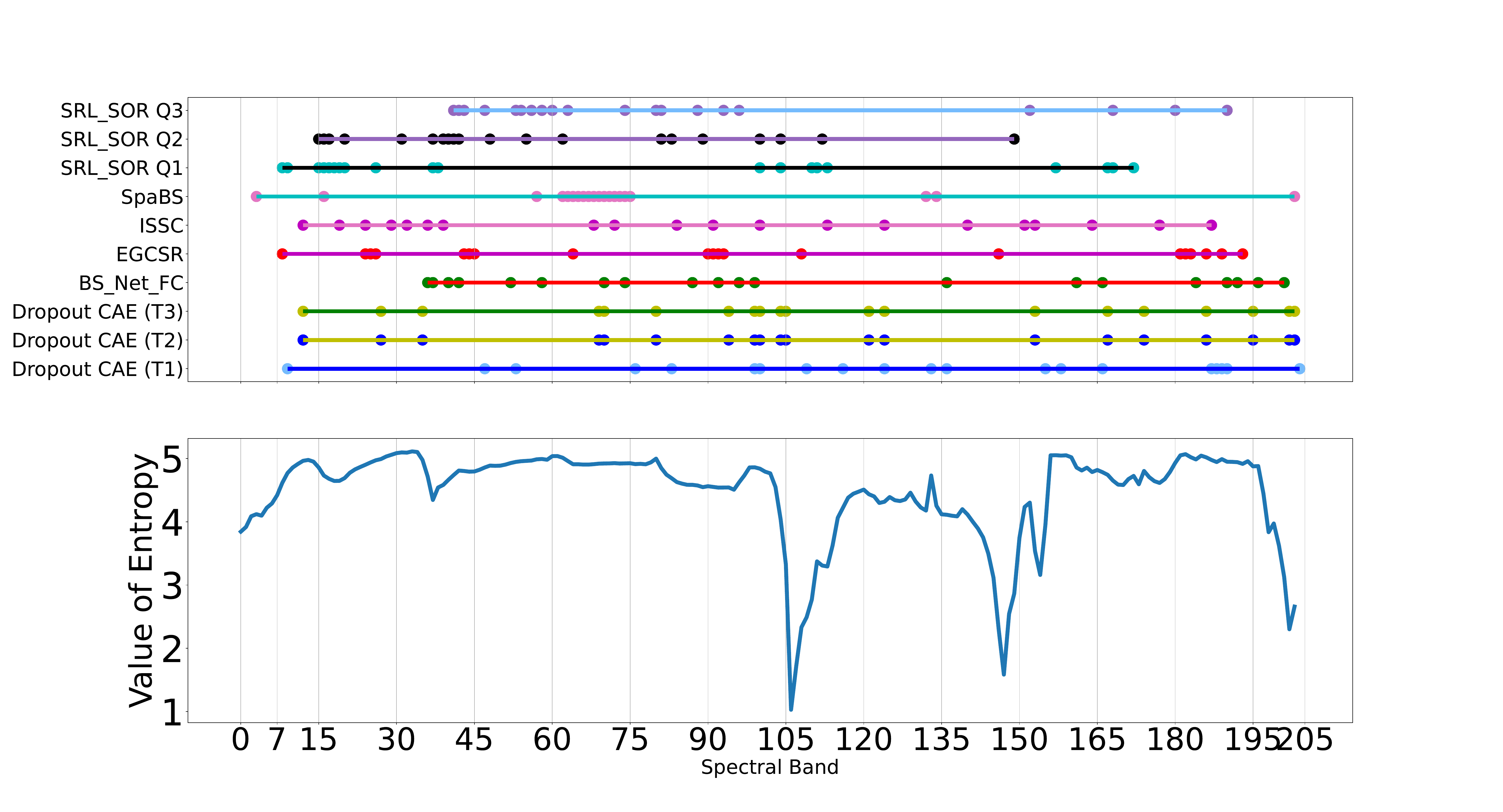}
  \caption{The selected subset with 20 bands and entropy values of each band on the Salinas scene}
  \label{fig:salinas_entropy}
\end{figure}

\begin{figure}[ht]
  \centering
  \includegraphics[width=1.1\columnwidth]{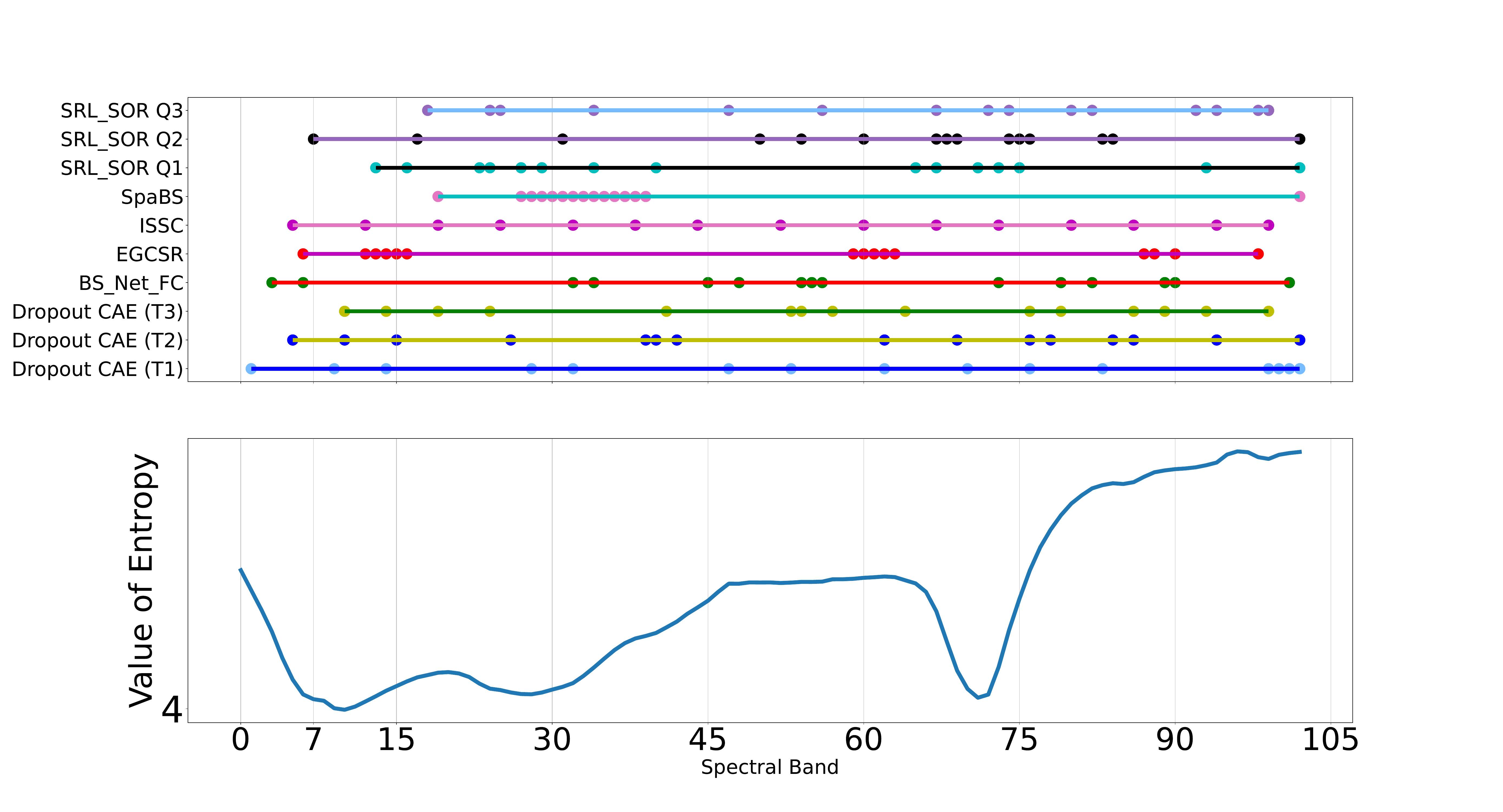}
  \caption{The selected subset with 15 bands and entropy values of each band on the PaivaU scene}
  \label{fig:paivau_entropy}
\end{figure}

\begin{figure}[ht]
  \centering
  \includegraphics[width=1.1\columnwidth]{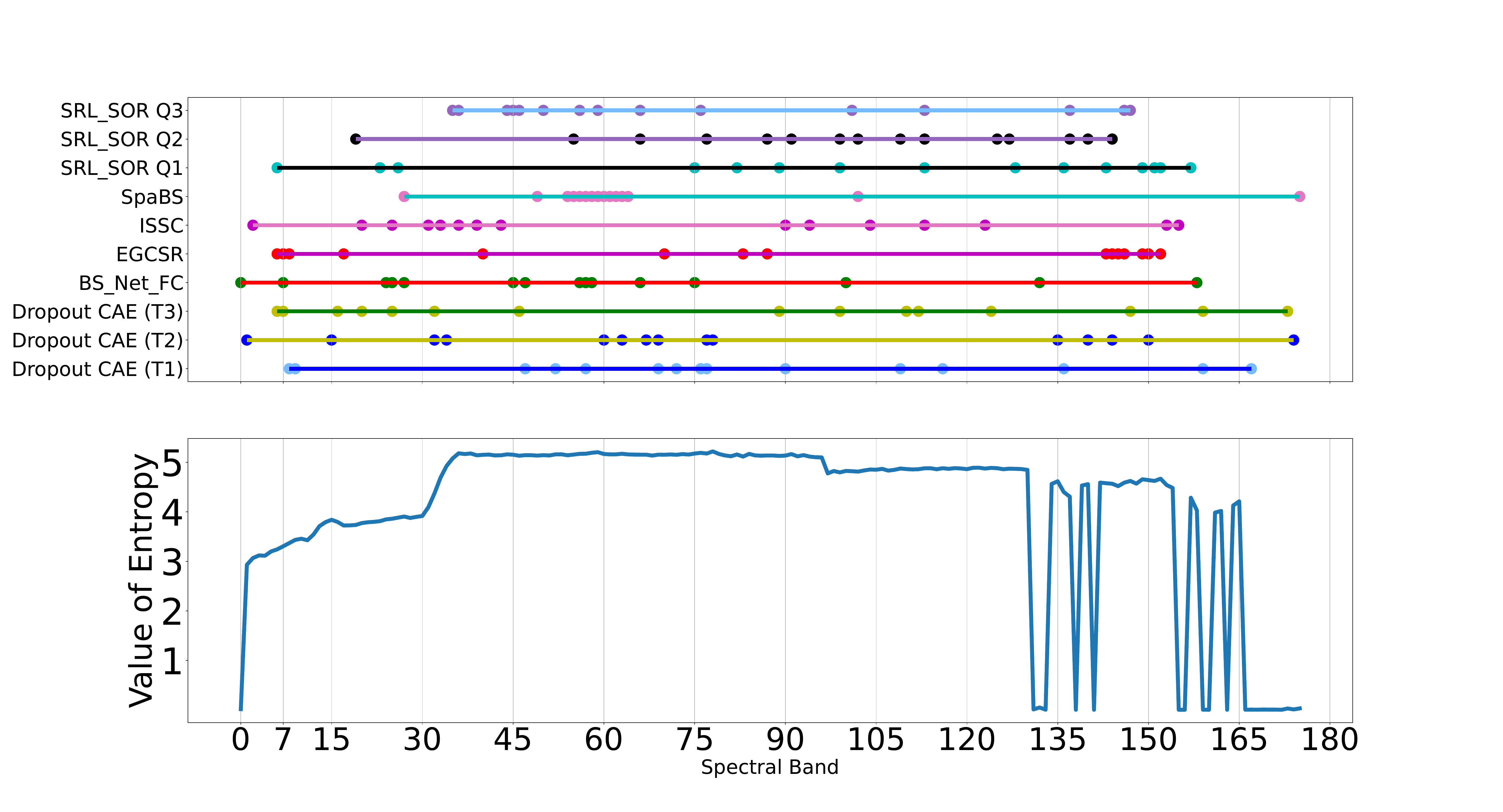}
  \caption{The selected subset with 15 bands and entropy values of each band on the KSC scene}
  \label{fig:ksc_entropy}
\end{figure}

We have used six different competing band selection methods for performance comparisons including SRL-SOA \cite{Ahishali2022SRL-SOA:SELECTION} with three Q values (polynomial approximation order), SpaBS \cite{Li2011SparseImages}, EGCSR \cite{Cai2021GraphImage}, ISSC \cite{Sun2015BandClassification}, and BS-Net-FC \cite{Cai2020BS-Nets:Image}. Moreover, three kinds of annealing schedules under the same condition are adapted for training to validate the effect of the temperature parameter $\tau$ with the dropout CAE model. We select 25 bands on Indian Pines scene, 20 bands on Sainas scene, 15 bands on PaviaU scene, and 15 bands on KSC scene with all methods. The top two results are highlighted in bold on each data scene. The overall quantitative results of the four data scenes are listed in Table. \ref{tab:results}. 

As shown in the Table. \ref{tab:results}, the dropout CAE has achieved the best and second best AA on the Indian Pines scene with $T_2$ and $T_1$ configurations as 0.7827 and 0.7786, respectively. Similarly, for the Salinas scene, the dropout CAE has obtained the best OA: 0.9614 and the second-best OA: 0.9611. Compared to other methods on the PaviaU, our proposed model has obtained the best AA and the second-best AA with $T_2$ and $T_1$ as 0.9926 and 0.9215 separately. In addition, the Dropout CAE ($T_2$) has achieved the best Kappa. For the KSC scene, the Dropout CAE ($T_2$) has the best AA. From Fig. \ref{fig:indian_entropy} to Fig. \ref{fig:ksc_entropy}, we demonstrate the distribution of the informative subset from various methods and entropy values of each band on each data scene for a visual comparison. In summary, the Dropout CAE has achieved robust and effective performance on different HSI scenes, compared to competing methods.

\section{Conclusion}
In this work, we propose a novel method named Dropout CAE to re-parameterize the discrete random variables for HSI band selection. We first utilize the variational dropout strategy to exploit the importance of each frequency band for HSI scenes. To bridge the gap between the discrete band information and the re-parameterization of the discrete random variables, we introduce the variational dropout strategy in binary concrete distribution enabling the Dropout CAE model to directly optimize the model weights together with learnable dropout rates. An extensive set of experiments on four data scenes shows that the proposed method outperforms the competing methods in HSI band selection task.


\bibliographystyle{IEEEbib}
\bibliography{references.bib}

\end{document}